\documentclass[10pt,twocolumn,letterpaper]{article}

\usepackage[pagenumbers]{iccv}

\usepackage{pifont}

\definecolor{iccvblue}{rgb}{0.21,0.49,0.74}
\usepackage[pagebackref,breaklinks,colorlinks,allcolors=iccvblue]{hyperref}
\usepackage{multirow}
\def\zl{\textcolor{black}}
\definecolor{DarkGreen}{HTML}{006400}
\definecolor{Darkred}{HTML}{7B241E}
\definecolor{light_red}{HTML}{F7E6E6}
\definecolor{light_green}{HTML}{E9EFEF}

\usepackage{color}
\usepackage{colortbl}
\definecolor{DarkGreen}{HTML}{006400}
\definecolor{Darkred}{HTML}{7B241E}
\definecolor{bluecolor}{HTML}{0071BC}
\definecolor{DeeperBlue}{HTML}{005B9A}
\definecolor{light_red}{HTML}{F7E6E6}
\definecolor{light_green}{HTML}{E9EFEF}
\usepackage{graphicx}
\usepackage{mathrsfs}
\makeatletter
  \newcommand\figcaption{\def\@captype{figure}\caption}
  \newcommand\tabcaption{\def\@captype{table}\caption}
\makeatother
\usepackage{makecell}

\usepackage{times}
\usepackage{epsfig}
\usepackage{amsmath}
\usepackage{amssymb}
\usepackage{wrapfig}
\usepackage{amsthm}

\usepackage{tabu}
\usepackage{bbm}
\definecolor{citecolor}{HTML}{2980b9}
\definecolor{linkcolor}{HTML}{c0392b}

\usepackage{float}
\usepackage[export]{adjustbox}
\usepackage{chngcntr}

\usepackage{natbib}  
\usepackage{caption} 
\usepackage{algorithm}
\usepackage{algorithmic}

\title{PiSA: A Self-Augmented Data Engine and Training Strategy\\for 3D Understanding with Large Models}

\author{Zilu Guo$^{1,2}$, Hongbin Lin$^{1,2}$, Zhihao Yuan$^{1,2}$, Chaoda Zheng$^{1,2}$, Pengshuo Qiu$^{3}$\\Dongzhi Jiang$^{3}$, Renrui Zhang$^{3}$, Chun-Mei Feng$^{4}$, Zhen Li$^{2,1}$\footnotemark[1]\vspace{0.2cm}\\
$^{1}$FNii-Shenzhen
$^{2}$SSE, CUHK-Shenzhen
$^{3}$CUHK\\
$^{4}$IHPC, A*STAR, Singapore\\
{\tt\small ziluguo1@link.cuhk.edu.cn}}

\begin{document}
\maketitle

\renewcommand{\thefootnote}{\fnsymbol{footnote}}
\footnotetext[1]{Corresponding Author.}

\begin{abstract}
3D Multimodal Large Language Models (MLLMs) have recently made substantial advancements. However, their potential remains untapped, primarily due to the limited quantity and suboptimal quality of 3D datasets. Current approaches attempt to transfer knowledge from 2D MLLMs to expand 3D instruction data, but still face modality and domain gaps.
To this end, we introduce \textit{\textbf{PiSA-Engine} (\textbf{P}o\textbf{i}nt-\textbf{S}elf-\textbf{A}ugmented-\textbf{Engine})}, a new framework for generating instruction point-language datasets enriched with 3D spatial semantics. We observe that existing 3D MLLMs offer a comprehensive understanding of point clouds for annotation, while 2D MLLMs excel at cross-validation by providing complementary information. By integrating holistic 2D and 3D insights from off-the-shelf MLLMs, PiSA-Engine enables a continuous cycle of high-quality data generation.
We select PointLLM as the baseline and adopt this co-evolution training framework to develop an enhanced 3D MLLM, termed \textit{\textbf{PointLLM-PiSA}}.
Additionally, we identify limitations in previous 3D benchmarks, which often feature coarse language captions and insufficient category diversity, resulting in inaccurate evaluations. To address this gap, we further introduce \textbf{\textit{PiSA-Bench}}, a comprehensive 3D benchmark covering six key aspects with detailed and diverse labels.
Experimental results demonstrate PointLLM-PiSA's \textit{state-of-the-art} performance in zero-shot 3D object captioning and generative classification on our PiSA-Bench, achieving significant improvements of 46.45\% \textbf{\textit{(+8.33\%)}} and 63.75\% \textbf{\textit{(+16.25\%)}}, respectively. 
We will release the code, datasets, and benchmark.
\end{abstract}    
\section{Introduction}
\label{sec:intro}
With recent advancements in 2D Multimodal Large Language Models (MLLMs)~\cite{openai2024gpt4technicalreport,liu2023llava,Qwen2VL,zhang2024llama,zong2024mova, jiang2024mmsearch,zhang2024mavis}, 3D MLLMs~\cite{3Dllm,qi2024shapellm,xu2023pointllm,guo2023point,tang2025exploring} have pushed the boundaries further by generating concise textual descriptions of 3D point clouds, enhancing their capability to integrate and interpret multiple modalities. This progress opens new avenues for leveraging these models to comprehend complex 3D data more effectively.

Current 3D works face three main problems as follows: \textit{\textbf{1) Limited Training Data.}}
Existing methods are constrained by small-scale training, as obtaining extensive 3D caption data is costly either by humans or by proprietary LLMs. 
\textit{\textbf{2) Poor Data Quality.} }
The quality of training data is often compromised when sourced through web scraping, introducing redundant and noisy data that can adversely affect the training process. 
This problem is first alleviated by Point-Bind LLM~\cite{guo2023point}, which adopts 2D MLLMs~\cite{han2023imagebind,gao2023llama} as a cross-modality bridge to entirely discard the need of 3D instruction-tuning data.
Further, PointLLM~\cite{xu2023pointllm} leverages training data derived from Cap3D~\cite{luo2023scalable} captions that are processed into instructions using GPT-4 without incorporating visual or point cloud context. This heterogeneity of visual and textual modalities during data processing significantly degraded the training data quality and, consequently, the model performance. Moreover, while manual annotation could potentially yield cleaner data, this approach would be prohibitively expensive and difficult to scale.
\textit{\textbf{3) Natural Defect of 2D Data.}} The prevailing training approach for 3D MLLMs relies heavily on using rendered images paired with textual descriptions leveraging the strengths of 2D MLLMs. Using 2D images as visual inputs frequently yields suboptimal results due to significant domain gaps between the point cloud and image modalities, often leading to interpretation errors. 
Moreover, even when multiple viewpoints are available, effectively synthesizing information from diverse perspectives remains a problem. 
Critical spatial information - including depth relationships, geometric properties and intricate spatial arrangements - is inevitably compromised during this transformation. The loss of 3D structural fidelity severely weakens the model's spatial reasoning abilities and amplifies hallucination tendencies, a well-documented challenge in MLLMs ~\cite{huang2023survey}.

To this end, we present \textit{\textbf{PiSA-Engine} (\textbf{P}o\textbf{i}nt-\textbf{S}elf-\textbf{A}ugmented\textbf{-Engine})}, a data generation engine and training strategy covering three stages, namely \textit{3D-space Data Annotation}, \textit{2D-space Data Refinement} and \textit{Itertive 3D Data Bootstrap.} 
PiSA-Engine leverages the complementary strengths of both 3D and 2D MLLMs. 
In the first stage, PiSA-Engine utilizes 3D MLLMs to extract essential 3D characteristics, including depth information, spatial relationships, and geometric properties of point clouds. 
In the second stage, we incorporate a powerful 2D MLLM as a verification module to ensure 2D description accuracy. Thanks to the rich 2D-text data pairs, 2D MLLMs have developed well, providing accurate descriptions of appearance, details, and other information. This stage examines 12 rendered views of the point cloud and refines visual attribute descriptions while preserving the original 3D spatial information, effectively serving as a cross-modal validation mechanism.
\begin{figure}[t!]
\centering
\includegraphics[width=0.45\textwidth]{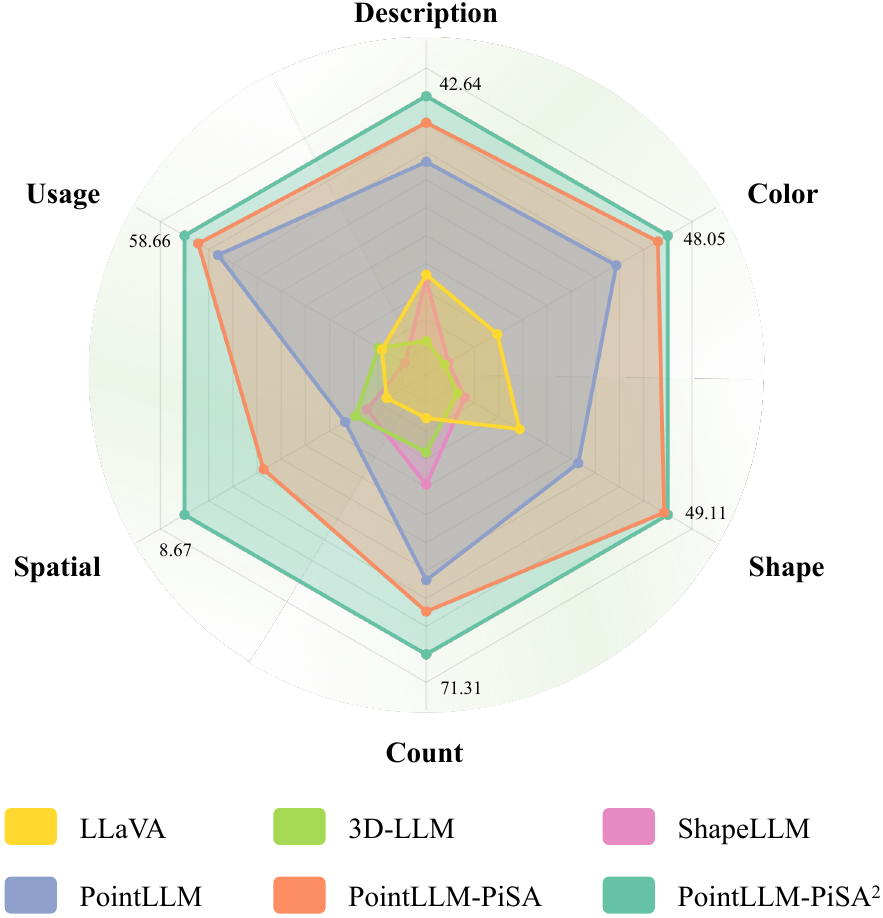}
\caption{
\textbf{Performance Comparison of Different Models on PiSA-Bench.} Considering the randomness of the generative task, we conduct five tests and take the average as the final result.}
\label{fig:benchmark}
\end{figure}
\begin{figure*}[h]
\centering
\includegraphics[width=\textwidth]{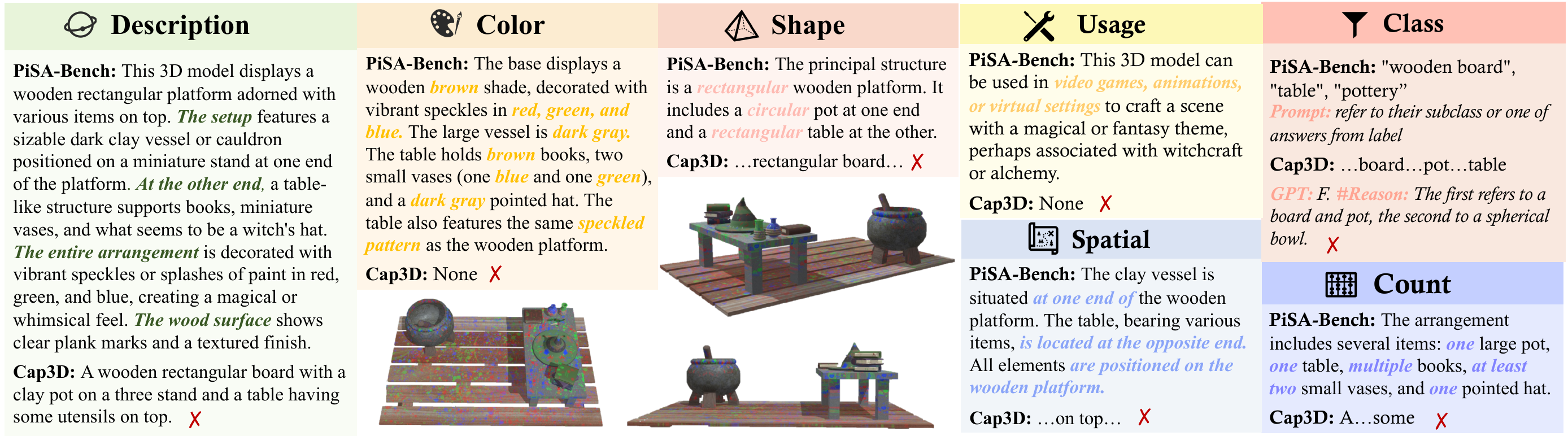}
\caption{
\textbf{Comparison between Previous and Proposed Benchmarks, PiSA-Bench.} In the testing set of PointLLM, each point cloud is paired with limited content, which is not comprehensive. However, the annotations in PiSA-Bench cover six comprehensive aspects. For the classification task, we provide class, subclass, and synonyms to avoid misjudgment.} 
\label{fig:3ddata}
\end{figure*}
In the third stage, we propose a co-evolution training strategy, where the model from the LOOP$_{t}$ phase is utilized to generate training data for the subsequent LOOP$_{t+1}$ phase.
We select PointLLM as the baseline and adopt this co-evolution training strategy to obtain an enhanced 3D MLLM, termed \textit{\textbf{PointLLM-PiSA}}. Under this setting, experiments have shown that the model in the LOOP$_{2}$ phase, termed PointLLM-PiSA$^{2}$, can achieve a 8.29\% improvement compared with PointLLM-PiSA.
Notably, tackling the limitations of 2D MLLMs, PointLLM-PiSA distinguishes itself by integrating semantic knowledge of 2D and 3D.

We further validate the effectiveness of PiSA-Engine by extending it to enhance existing 3D understanding tasks. For example, with the improved dataset from PiSA-Engine, we observe that the
\textit{top-1} accuracy for zero-shot classification on ScanObjectNN increase to 64.42\%.

Finally, we find that prevail 3D benchmarks~\cite{deitke2023objaverse,wu20153d,luo2023scalable} are not comprehensive, most of which only cover short and rough annotations. There are several defects in the current dataset, as shown in Figure~\ref{fig:comparison}. To remedy these defects, we introduce \textbf{\textit{PiSA-Bench}}, a comprehensive 3D benchmark. As shown in Figure~\ref{fig:benchmark} and Figure~\ref{fig:3ddata}, PiSA-Bench can thoroughly evaluate each 3D MLLM from 6 aspects. 

We summarize the key contributions as follows:

\begin{itemize}

    \item To the best of our knowledge, we are the \textit{\textbf{{first}}} to introduce an Automatic 3D Data Annotation Engine and Iterative Training Strategy, named PiSA-Engine, which can scale existing point-text pair datasets.
    \zl{Meanwhile, compared to 2D captions, our PiSA-Engine further improves data quality.}
    
    \item We introduce PointLLM-PiSA, which is enhanced by PiSA-Engine. Experiments show that PointLLM-PiSA has achieved remarkable improvement across tasks, \ie, achieving a \textbf{\textit{+8.29\%}} gain, raising the accuracy to 46.41\%, in 3D object captioning on PiSA-Bench and a new \textit{state-of-the-art} of an average accuracy 61.25\% \textbf{\textit{(+16.25\%)}} on generative 3D object classification. 
    
    \item  
    We propose PiSA-Bench to evaluate 3D object captioning and generative classification tasks jointly. It covers general description, color, shape, count, spatial, usage, hallucination, and class aspects. We plan to release it for future research development.

\end{itemize}   
\section{Related Work}
\label{sec:related}

\vspace{0.1cm}

We first review the literature on 3D Multimodal Large Language Models, and then discuss 3D Object Understanding with Language methods, and lastly we explore the research on Self-Augmented Large Language Models. 

\vspace{0.1cm}
\subsection{3D Multimodal Large Language Models}

In the realm of 3D multimodal large language models, Point-Bind LLM~\cite{guo2023point}, as preliminary work, utilizes a 3D multi-modal encoder, Point-Bind, to construct 3D MLLM without using 3D instruction data, while facing challenges with hallucinations due to its dependency on retrieval methods. 
Then, PointLLM~\cite{xu2023pointllm} utilizes automated instruction data generation; however, its instruction generation process in GPT relies solely on text, omitting 2D and 3D spatial context, which increases susceptibility to errors and hallucinations. ShapeLLM~\cite{qi2024shapellm} and GPT4Point~\cite{GPT4Point} incorporate rendered images in data generation, yet they lack true 3D information and may introduce inaccuracies due to inconsistencies in 2D viewpoints. Research~\cite{li2023blip,gong2023multimodal, radford2021learningtransferablevisualmodels, jia2021scaling} supports 3D-LLM~\cite{deitke2023objaverse}, which leverages multi-view image features from object renderings and extensive image-language datasets to enhance performance on downstream tasks. 
These models typically rely on 2D MLLMs for feature extraction, helping bridge language models toward a more comprehensive 3D understanding.

\vspace{0.1cm}
\subsection{3D Object Understanding with Language}

Through the integration of visual projections from point clouds, advanced models such as PointCLIP~\cite{zhang2021pointclippointcloudunderstanding}, PointCLIPv2~\cite{zhu2023pointclipv2promptingclip}, and CLIP2Point~\cite{huang2023clip2pointtransferclippoint} are enhancing 3D recognition capabilities by leveraging established CLIP~\cite{radford2021learningtransferablevisualmodels} frameworks. ULIP~\cite{xue2022ulip} was the first to fuse 3D point cloud data with multimodal inputs from images and text, significantly advancing 3D representation learning and addressing some of the challenges posed by limited 3D-text data. 

Further progress is exemplified by models like JM3D~\cite{Wang_2023}, CG3D~\cite{hegde2023clipgoes3dleveraging}, and Uni3D~\cite{zhou2023uni3d}, which refine their algorithms by aligning point cloud encoders with CLIP embeddings using a triplet structure incorporating point clouds, images, and text. These approaches largely depend on internet-sourced data, such as Objaverse~\cite{deitke2023objaverse}, which often includes short, noisy text annotations. Additionally, human annotations frequently fall short of conveying the full semantic richness present in 3D data.

\begin{figure*}[t!]
    \centering
    \includegraphics[width=\linewidth]{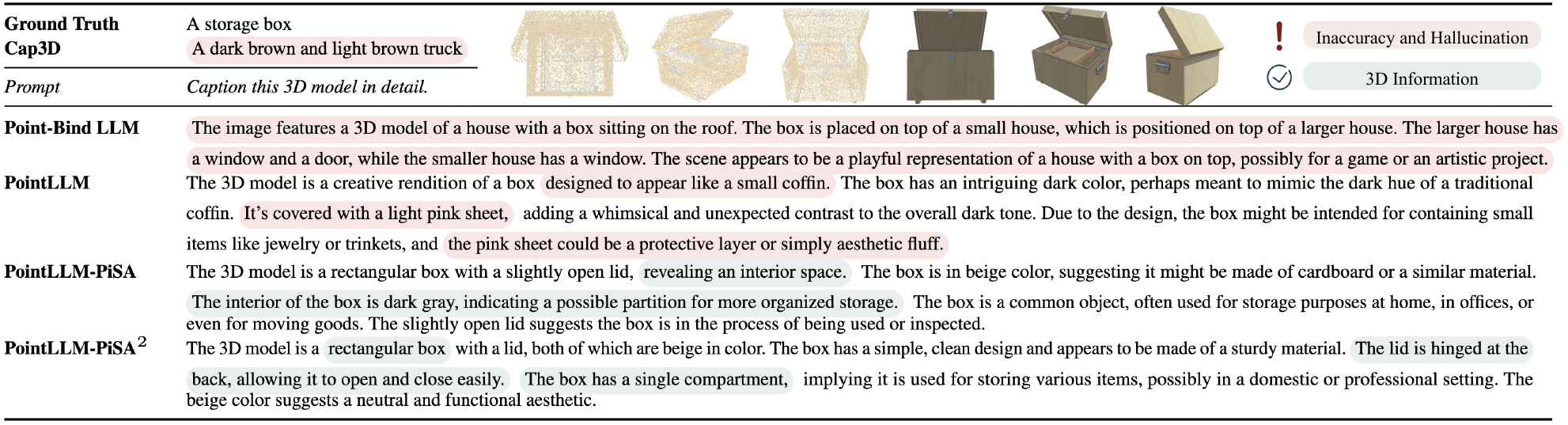}
    \caption{\textbf{Sample Comparisons.} We showcase one example of \textbf{Inaccuracy and Hallucination} (highlighted in red) and \textbf{3D information including depth, spatial, relative position and geometric Information} (highlighted in green) within existing 3D MLLMs~\cite{guo2023point,xu2023pointllm} and ours. These samples show our method produces more accurate and detailed results than the baseline and human-annotated ground truths.}
    \label{fig:comparison}
\end{figure*}

Models like OpenShape~\cite{liu2023openshape} and ULIP-2~\cite{xue2023ulip2} utilize image-captioning algorithms to generate synthetic textual data, thus enhancing the quality of the triplet data. While OpenShape leverages multimodal language models, it restricts captioning to 2D images from fixed perspectives, potentially overlooking essential 3D attributes like depth and spatial relationships. Similarly, ULIP-2 employs BILP-2~\cite{li2023blip} to generate detailed captions but limits the textual input to the top-k CLIP-ranked captions, which may omit critical semantic information necessary for precise 3D feature alignment during training. Consequently, these existing methods still fall short of achieving a comprehensive understanding of native 3D point clouds.

\subsection{Self-Augmented Large Language Models}
Self-augmented large language models (LLMs) represent a transformative advancement in enhancing the efficiency and capability of LLMs through innovative data curation and training methodologies. Prior research~\cite{zhou2023lima,li2024quantity} emphasizes the importance of dataset quality over sheer quantity during the instruction fine-tuning stage.
Additionally, synthetic data generation has gained prominence, leveraging the generative capability of LLMs or closed-model outputs to augment datasets. Techniques such as rephrasing, question-answer synthesis, and iterative improvement processes demonstrate the potential of dynamic self-enhancement frameworks like I-SHEEP~\cite{alemohammad2024selfimprovingdiffusionmodelssynthetic}, transitioning beyond static, one-time data augmentation to continuous model improvement~\cite{wang2023selfinstruct,sun2023conifer}.

Iterative enhancement methods, often involving collaboration with strong models such as GPT-4 and Claude, detect and refine errors in LLM outputs to ensure safety and reliability. Notable examples, such as IterAlign~\cite{chen2024iteralign}, demonstrate how iterative reinforcement learning from human feedback (RLHF)~\cite{yuan2024rlhf} aligns models with human preferences and prevents model collapse. Furthermore, the iterative enhancement methods also witness advancements in the multimodal domain~\cite{deng2024efficient,NEURIPS2024_ed45d6a0,alemohammad2024self,jiang2024comat,yuan2024self,zong2024easyref}.
For example, in the domain of 2D MLLM, \cite{NEURIPS2024_ed45d6a0} integrate self-generated data to augment reasoning~\cite{zhang2024mathverse, jiang2025mme} and perception~\cite{goyal2017making, fu2024blink} capabilities, achieving state-of-the-art performance without reliance on commercial APIs~\cite{radford2021clip,peng2023multimodal}.

These advancements underscore the potential of self-augmented paradigms in optimizing LLM capabilities through scalable, iterative, and efficient strategies. This line of work has seen significant progress in the 2D domain but remains largely unexplored in the 3D domain. PiSA adopts the proven paradigm of the 2D domain and extends it to tackle key challenges in the 3D domain, specifically addressing issues of data scarcity and model generalization.
\section{PiSA-Engine}

\begin{figure}[t!]
\centering
\includegraphics[width=0.48\textwidth]{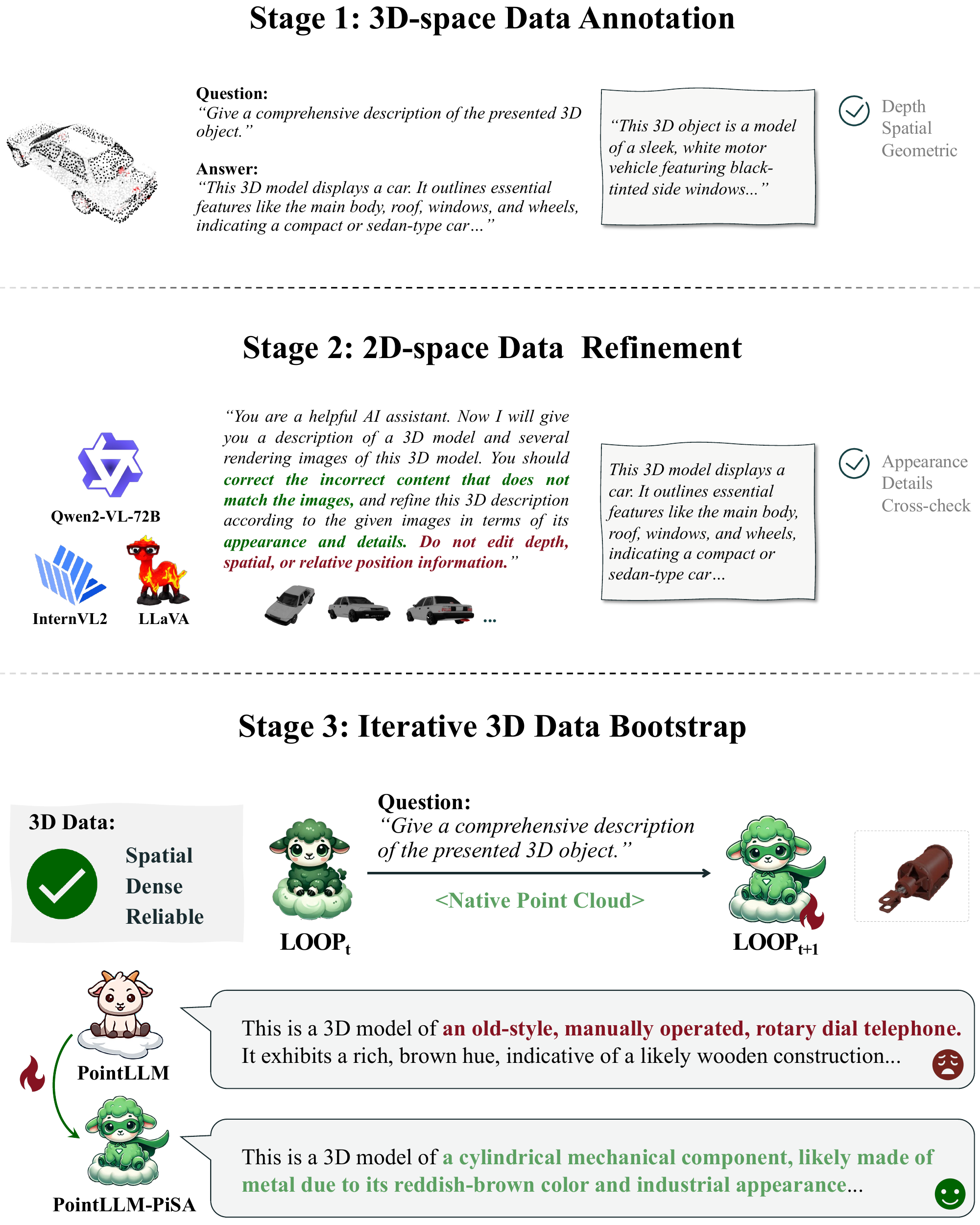}
   \caption{\textbf{The Overview of the PiSA-Engine.} After enhanced by data generated by the PiSA-Engine during the LOOP$_{t}$ phase, the PointLLM-PiSA can produce higher-quality outputs, 
   which can inject more precise 3D knowledge into the training data for LOOP$_{t+1}$ phase.
   The above shows this recursive training strategy.
   }
\label{fig:main}
\end{figure}
Figure~\ref{fig:comparison} demonstrates that captions from Cap3D~\cite{deitke2023objaverse,luo2023scalable,luo2024view} and Point-Bind LLM~\cite{guo2023point} can contain significant inaccuracies. In contrast, the 3D captions produced by PointLLM~\cite{xu2023pointllm} accurately describe the object but lack details on its relative position, spatial relationships, geometric structure, and depth. Previous approaches have either employed human annotators to write dense captions - an approach that is resource-intensive and costly - or utilized commercial APIs for 2D descriptions based on rendered images, which often lose essential 3D information. However, detailed and comprehensive 3D captions are valuable for training 3D language models, as we will discuss later. For example, 2D MLLMs are unable to perceive occlusions (\eg, one object behind another), and, as shown in Figure~\ref{fig:comparison}, this limitation can lead to critical errors due to the loss of such spatial information.

As illustrated in Figure~\ref{fig:main}, our proposed method, PiSA-Engine, comprises three main stages: \textit{3D-space Data Annotation}, \textit{2D-space Data Refinement} and \textit{Itertive 3D Data Bootstrap.} We begin by introducing these stages and then demonstrate our proposed PointLLM-PiSA. Next, we clarify the details and production process of PiSA-Bench.

\subsection{3D-space Data Annotation}
Current 3D MLLMs mostly follow the render-annotation paradigm for 3D data annotation, resulting in two problems.
Firstly, the rendering process causes information loss, making it impossible to preserve the native geometric properties of the point cloud.
Secondly, view differences can lead to contradictions in the semantic content, especially in complex point cloud structures.
Therefore, we adopt 3D MLLMs as the first step to solve these problems.
By leveraging the capabilities of the already intelligent PointLLM at the first stage, we can handle intensive relabeling tasks efficiently. This approach ensures that captions incorporate essential depth, spatial, and geometric information. At this stage, prompts should be straightforward, given PointLLM’s limitations with complex prompts. For instance, a suitable prompt might be: \textit{``Caption this 3D model in detail, describing its depth, spatial, and geometric information".}

\subsection{2D-space Data Refinement}
To avoid inevitable mistakes, we leverage 2D MLLMs with image information to cross-check and refine the annotation; meanwhile, we keep the 3D information intact, as shown in Figure~\ref{fig:main}.
In the render-annotation paradigm, previous methods always caption images of multiple views (6-12), which is costly. However, we only refine one captioning annotation for each data pair, namely \textbf{\textit{2D-space Data Refinement}}.

In this stage, we employ Qwen2-VL-72b~\cite{Qwen2VL} to refine 3D captions based on 12 rendered images, 
enhancing the semantic richness and aiding in the alignment of features across the subsequent three modalities. The prompt used is:\textit{``You are a helpful AI assistant. Now I will give you a description of a 3D model and several rendering images of this 3D model. You should correct the incorrect content that does not match the images, and refine this 3D description according to the given images in terms of its appearance and details. Do not edit depth, spatial, or relative position information."} Our experimental results indicate that incorporating the information of rendered images makes the caption more accurate and fixes errors.

\subsection{Iterative 3D Data Bootstrap}
After thoroughly examining the challenges of 3D MLLMs, we introduce a novel strategy of iteration to tackle the issue of data scarcity, namely \textbf{\textit{Itertive 3D Data Bootstrap}}. This method aims to address the data annotation challenge through a self-play model, which recursively improves its understanding of native point cloud data. Through iterative training, we propose a \textbf{data-model co-evolution framework} that forms a positive feedback loop where the data enhances the model, and then the model refines the data. As a plug-and-play co-evolution framework, it can be applied to any 3D MLLMs.

\begin{figure}[t!]
\centering
\includegraphics[width=0.48\textwidth]{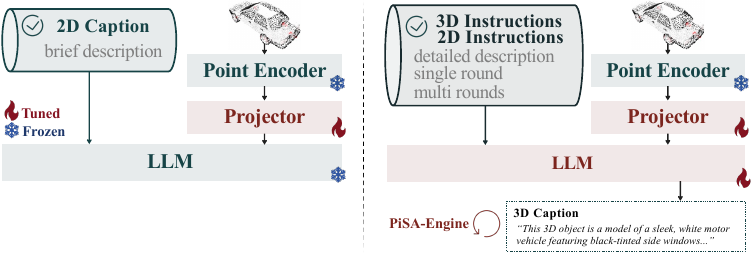}
\caption{\textbf{Training Pipeline of PointLLM-PiSA}. We integrate 2D and 3D semantic knowledge through multi-modal data and end-to-end training for enhanced point cloud understanding.}
\label{train}
\end{figure}

\subsection{PointLLM-PiSA}
\label{sec:self-augment}

Given that we aim to leverage our newly-created datasets from PiSA-Engine to enhance the ability of 3D MLLMs,  
PointLLM-PiSA is trained on a combined dataset consisting of original instruction data and additional data generated by the PiSA-Engine, allowing the model to leverage multimodal semantic insights. 
Specifically, we train PointLLM-PiSA by minimizing the negative log-likelihood of the text token at each position, following the components and the training setting of PointLLM~\cite{xu2023pointllm}. The end-to-end nature of this training approach enables the model to effectively integrate point cloud and text modalities.
Overall, the training scheme of PointLLM-PiSA is as follows:

\begin{equation}
    \mathcal{L} = -\sum_{i = 1}^{L} \log p(\tilde{z}_i | \tilde{\textbf{z}}_{<i}, \textbf{z}_{ins}; \Theta),
\end{equation}
where $\tilde{z}_i$ are the tokens from the responses of the model including \texttt{</s>} and $i$ indexes each token in the response. 
$L$, $\textbf{z}_{ins}$, $\Theta$ and denotes the length of the response, instructions tokens and model parameters, respectively.

Given a point cloud with coordinates of specific points or regions, along with descriptive instructions, we tokenize and encode these inputs to generate point cloud embeddings, denoted as $\textbf{z}_p$, and text embeddings, $\textbf{z}_t$. We then align the point features with the word embedding space within the language model. Finally, we integrate the point cloud and language representations, forwarding them jointly into the LLM to enable the understanding of native point cloud, as depicted in Figure~\ref{train}.

\begin{table}[t]
\centering
\setlength\tabcolsep{2.5pt}
\scalebox{0.6}{
\begin{tabular}{@{}l@{\hspace{0.13em}}c@{\hspace{-0.1em}}ccccc}
\toprule
Model & Input & Objaverse (I) & Objaverse (C) & Average \\ \midrule
LLaVA-7B~\cite{liu2023llava} & Single-V. Img. & 31.50 & 33.50 & 32.50 \\
LLaVA-13B~\cite{liu2023llava} & Single-V. Img. & 34.00 & 36.50 & 35.25 \\ 
3D-LLM~\cite{hong20233dllminjecting3dworld} & 3D Obj. + Mul.-V. Img. & 32.00 & 27.50 & 29.75 \\
ShapeLLM-7B~\cite{qi2024shapellm} & 3D Point Cloud & 41.00 & 34.00 & 37.50 \\
ShapeLLM-13B~\cite{qi2024shapellm} & 3D Point Cloud & 44.50 & 43.00 & 43.75 \\
Point-Bind LLM-7B~\cite{guo2023point} & 3D Point Cloud & 1.00 & 0.50 & 0.75 \\ \midrule
PointLLM-7B~\cite{xu2023pointllm} & 3D Point Cloud & 39.00 & 36.00 & 37.50 \\
\rowcolor{light_green}\textbf{PointLLM-PiSA-7B}& 3D Point Cloud & 45.00 & 46.50 & \textbf{45.75} \\
\rowcolor{light_green}\textbf{PointLLM-PiSA$^{2}$-7B}& 3D Point Cloud & 46.00 & 48.00 & \textbf{47.00} \\

\rowcolor{light_green}\textbf{PointLLM-PiSA$^{3}$-7B} & 3D Point Cloud&46.50 & 47.50 & \textbf{47.00}\\

PointLLM-13B~\cite{xu2023pointllm} & 3D Point Cloud & 37.50 & 40.0 & 38.75 \\ 
\rowcolor{light_green}\textbf{PointLLM-PiSA-13B}& 3D Point Cloud & 51.50 & 49.50 & \textbf{50.50} \\
\rowcolor{light_green}\textbf{PointLLM-PiSA-13B$^{*}$}& 3D Point Cloud & 53.50 & 50.50 & \textbf{52.00} \\
\toprule
Model & Input & PiSA-Bench (I) & PiSA-Bench (C) & Average \\ \midrule
3D-LLM & 3D Obj. + Mul.-V. Img. & 20.00 &32.50& 26.75 \\ 
ShapeLLM-7B & 3D Point Cloud & 27.50&37.50&32.50 \\
ShapeLLM-13B & 3D Point Cloud & 55.00&42.50&48.75 \\
Point-Bind LLM-7B & 3D Point Cloud & 7.50 & 5.00 & 6.25 \\ \midrule
PointLLM-7B & 3D Point Cloud & 47.50&47.50&47.50  \\
\rowcolor{light_green}\textbf{PointLLM-PiSA-7B} & 3D Point Cloud & 60.00  &62.50 & \textbf{61.25} \textcolor{DarkGreen}{\small ($\uparrow 13.75$)} \\
\rowcolor{light_green}\textbf{PointLLM-PiSA$^{2}$-7B} & 3D Point Cloud & 62.50  &62.50& \textbf{62.50} \textcolor{DarkGreen}{\small ($\uparrow 15.00$)} \\
\rowcolor{light_green}\textbf{PointLLM-PiSA$^{3}$-7B} & 3D Point Cloud & 60.00 &67.50 & \textbf{63.75} \textcolor{DarkGreen}{\small ($\uparrow 16.25$)} \\

PointLLM-13B & 3D Point Cloud & 52.50&55.00&53.75 \\ 
\rowcolor{light_green}\textbf{PointLLM-PiSA-13B} & 3D Point Cloud & 55.00  & 57.50  & \textbf{56.25} \textcolor{DarkGreen}{\small ($\uparrow 2.50$)} \\ 
\rowcolor{light_green}\textbf{PointLLM-PiSA-13B$^{*}$} & 3D Point Cloud & 67.50 & 65.00  & \textbf{66.25} \textcolor{DarkGreen}{\small ($\uparrow 12.50$)} \\ \bottomrule
\end{tabular}}
\caption{\textbf{Zero-shot Generative 3D Object Classification Results on Objaverse and PiSA-Bench.} The results show the classification accuracy under the \textbf{I}nstruction-typed (I) prompt ``What is this?" and the \textbf{C}ompletion-typed (C) prompt ``This is an object of ". \textit{(We obtained the above experimental results via open-source code, and the same applies below. $^{*}$indicates those models trained with 182K data.)}}
\label{table:gen_cls}
\end{table}

\section{PiSA-Bench: Comprehensive 3D Benchmark}

Experimental results demonstrate that our model outperforms PointLLM \cite{xu2023pointllm} across traditional, GPT-4o, and human evaluation. Furthermore, we identify several issues with previous 3D benchmarks. First, the 3D annotation data are coarse and lack details.
Secondly, in generative classification tasks, we observe that GPT-4o sometimes incorrectly scores similar outputs, such as labeling a model output of ``mug" as incorrect compared to the ground truth ``cup", which we consider unreasonable. Thirdly, as shown in Figure~\ref{fig:comparison}, the testing set used by PointLLM contains significant errors, such as mislabeling a cardboard box as a truck, which introduces inaccuracies in evaluating generative classification and captioning tasks. To address these limitations, we propose a more accurate and comprehensive 3D benchmark, PiSA-Bench, for evaluating captioning and classification tasks on native point clouds.
 
As shown in Figure~\ref{fig:3ddata}, we have meticulously designed 240 point-text pairs in PiSA-Bench, ranging in 6 aspects $\times$ 
40 point cloud objects.
Specifically, our observations show that MLLMs often produce erroneous claims about the point cloud in some aspects. Consequently, we design PiSA-Bench around these aspects: description, color, shape, count, spatial relations and usage. Detailed scoring criteria for each of these six aspects can be found in Appendix \emph{\textbf{C}}.

To ensure fairness and comprehensiveness, we utilize GPT-4 to rephrase 3D captions generated by existing 3D MLLMs across six key aspects.
The rephrasing prompt provided for GPT-4 is: ``\textit{Rephrase this description of a 3D object using totally different vocabulary and sentence structure. Keep the original information accurate, do not add more information".} To verify accuracy, experts then review these rephrased captions using visualizations from the Objaverse dataset's official website \cite{deitke2023objaverse}. During the human review, we ensure the six-aspect descriptions are free from complex vocabulary, compound nouns, and overly intricate sentence structures, eliminating excessive interpretation. After processing by GPT-4 and expert review, the resulting ground-truth descriptions comprehensively cover all key characteristics of each point cloud across six key aspects.

To address the second benchmark challenge for PointLLM \cite{xu2023pointllm}, we use GPT-4 to generate 3-5 synonyms as ground-truth labels for classification. The testing prompt is set as: \textit{``Remember the answer from model refers to one of the answers from the label, even the answer from model refers to the subclass of one of the answers from the label, you should respond `T'".}

\section{Experiments}

In this section, we present experimental results and analysis on a range of downstream tasks, complemented by ablation studies to assess the effectiveness of our method.

\subsection{Experimental Settings}
\label{settings}
We train PointLLM-PiSA for three epochs with a batch size of 16 and adopt AdamW as the optimizer with a learning rate of 2\text{e}{-4}. All experiments are conducted on 8 A100 GPUs. Note that ``GPT-4o" in this paper refers specifically to OpenAI’s GPT-4o model. 

We utilize PiSA-Engine to generate 62K instruction data from Cap3D~\cite{deitke2023objaverse,luo2023scalable,luo2024view} as our training dataset. This augmented dataset, combined with the original 70K PointLLM instruction training dataset, is used to enhance PointLLM, therefore we get PointLLM-PiSA. Although our processed data account for only \textbf{13.28\%} of the overall training dataset, it provides a significant improvement in performance.
Subsequently, we leverage PointLLM-PiSA to generate an additional 50K higher-quality instruction data samples to serve as training data for the next iteration, termed PointLLM-PiSA$^{2}$. Following this experimental logic, proceeding to the next iteration, we obtained PointLLM-PiSA$^{3}$. PointLLM-PiSA-13B$^{*}$ means that in the first iteration, we enhance PointLLM-13B with a total of 182K instruction data. Detailed data components can be found in Appendix \emph{\textbf{A}}.

\subsection{Generative 3D Object Classification}

\begin{table*}[t!]
\centering
\setlength\tabcolsep{7pt}
\scalebox{0.95}{
\begin{tabular}{@{}l@{\hspace{0.7em}}c@{\hspace{-0.05em}}cccccccc@{}}
\toprule
Model & Input & Desc. & Color & Shape & Count & Spatial & Usage & AVG \\ \midrule
LLaVA-7B~\cite{liu2023llava} & Single-V. Img. & 23.04 & 20.50 & 28.21 & 40.58 & 1.42 & 22.18 & 22.66 \\
LLaVA-13B~\cite{liu2023llava} & Single-V. Img. & 24.27 & 18.97 & 25.16 & 45.42 & 1.89 & 23.03 & 23.12 \\ 
\midrule
3D-LLM~\cite{hong20233dllminjecting3dworld} & 3D Obj. + Mul.-V. Img. & 15.75 & 11.96 & 19.39 & 45.03 & 2.54 & 22.80 & 19.58 \\
ShapeLLM-7B~\cite{qi2024shapellm} &3D Point Cloud& 22.47 & 12.57 & 20.45 & 49.21 & 2.13 & 17.94 & 20.80 \\
ShapeLLM-13B& 3D Point Cloud  & 25.13 & 21.67 & 28.28 & 50.90 & 0.13 & 19.49 & 24.27 \\
Point-Bind LLM-7B~\cite{guo2023point} & 3D Point Cloud & 1.54 & 0.00 & 2.31 & 3.08 & 0.00 & 1.92 & 1.48 \\ \midrule
PointLLM-7B & 3D Point Cloud & 35.42 & 39.74 & 36.47 & 61.67 & 2.91 & 52.48 & 38.12 \\

 \rowcolor{light_green}\textbf{PointLLM-PiSA-7B} & 3D Point Cloud & \textbf{39.72} \textcolor{DarkGreen}{($\uparrow 4.30$)}&46.44&48.31& 65.74 &5.84&56.15&\textbf{43.70} 
 \textcolor{DarkGreen}{($\uparrow 5.58$)}\\ 
\rowcolor{light_green}\textbf{PointLLM-PiSA$^{2}$-7B} & 3D Point Cloud& \textbf{42.64} \textcolor{DarkGreen}{($\uparrow 7.22$)} & 48.05 & 49.11 & 71.31 & 8.67 & 58.66 & \textbf{46.41} \textcolor{DarkGreen}{($\uparrow 8.29$)} \\
\rowcolor{light_green}\textbf{PointLLM-PiSA$^{3}$-7B} & 3D Point Cloud& \textbf{42.76} \textcolor{DarkGreen}{($\uparrow 7.34$)} & 48.16 & 48.88 & 71.05 & 8.78 & 59.07 & \textbf{46.45} \textcolor{DarkGreen}{($\uparrow 8.33$)} \\

PointLLM-13B & 3D Point Cloud & 37.44 & 42.69 & 46.15 & 62.18 & 8.33 & 49.96 & 41.13 \\ 
\rowcolor{light_green}\textbf{PointLLM-PiSA-13B} & 3D Point Cloud & \textbf{42.95} \textcolor{DarkGreen}{($\uparrow 5.51$)}& 49.87&52.31&73.08&9.62&53.64&\textbf{46.91} \textcolor{DarkGreen}{($\uparrow 5.78$)}\\
\bottomrule
\end{tabular}
}
\caption{\textbf{Zero-shot 3D Object Captioning Results on PiSA-Bench.} Models are evaluated using GPT-4o, with the prompt: \textit{``Caption this 3D model in detail.} Considering the randomness of the generative task, we conduct five tests and take the average as the final result.}
\label{table:pisa-bench_cap}
\end{table*}

\begin{table}[t!]
\centering
\setlength\tabcolsep{1.8pt}
\scalebox{0.6}{
\begin{tabular}{@{}l@{\hspace{0.35em}}cccccccc}
    \toprule
Model  & GPT-4o & Sentence-BERT & SimCSE & BLEU-1 & ROUGE-L & METEOR & Corr.$\uparrow$ \\ 
    \midrule
ShapeLLM-7B & 28.04 & 41.61 & 42.18 & 6.60 & 8.40 & 13.25 & 2.49 \\ 
Point-Bind LLM-7B & 2.10 & 26.54 & 24.66 & 2.59 & 6.03 & 9.48 & 0.14 \\ 
PointLLM-7B & 28.41 & 46.59 & 47.18 & 3.55 & 6.77 & 10.97 & 4.56 \\ 
\rowcolor{light_green}\textbf{PointLLM-PiSA-7B} & \textbf{44.87} & \textbf{48.21} & \textbf{48.38} & \textbf{3.81} & \textbf{7.29} & \textbf{12.32} & \textbf{5.78} \\
\rowcolor{light_green}\textbf{PointLLM-PiSA$^{2}$-7B} & \textbf{47.91} & \textbf{48.66} & \textbf{49.04} & 3.78 & 7.11 & \textbf{12.46} & \textbf{5.94} \\
    \midrule
ShapeLLM-13B & 32.52 & 42.12 & 43.03 & 7.68 & 9.52 & 14.27 & 3.44 \\
PointLLM-13B & 33.69 & 48.90 & 50.21 & 3.75 & 7.17 & 12.38 & 4.76 \\ 
\rowcolor{light_green}\textbf{PointLLM-PiSA-13B} & \textbf{47.45} & \textbf{49.78} & \textbf{50.09} & \textbf{4.00} & \textbf{7.62} & \textbf{13.36} & \textbf{6.32} \\ 
    \bottomrule
\end{tabular}}
\caption{\textbf{Zero-shot 3D Object Captioning Results on Objaverse.} Models are evaluated using human evaluation, GPT-4o evaluation, and traditional metrics. Compared to PointLLM, which uses the 2023 version of GPT-4, we use the latest GPT-4o as the evaluation. We retest these methods since its scoring criteria differ from GPT-4. Considering the randomness of the generative task, we conduct five tests and take the average as the final result.}
\label{tab:obj_caption}
\end{table}
As shown in Table~\ref{table:gen_cls}, we present a comprehensive comparison of various models on zero-shot generative 3D object classification tasks. Each model was evaluated using two prompt types: an Instruction-style (I) prompt, ``What is this?" and a Completion-style (C) prompt, ``This is an object of." Our experimental results demonstrate that PointLLM-PiSA consistently outperforms others across both prompt types on Objaverse test datasets, underscoring PointLLM-PiSA’s strong generalization capabilities. On PiSA-Bench, our method also performs best with an improvement of \textbf{\textit{(+13.75\%)}} on average compared with the baseline model PointLLM. After three iterations, performance continues to improve, reaching an average score of 63.75\%.

It is necessary to mention that the 13B model requires more high-quality data to prevent overfitting. To address this issue, we annotate additional samples with PiSA-Engine and enhanced PointLLM-PiSA on the expanded 182K dataset. As shown in Table~\ref{table:gen_cls}, PointLLM-PiSA-13B$^{*}$ demonstrates significantly improved performance.

\subsection{3D Object Captioning}
To assess the effectiveness of the zero-shot captioning task, we utilize a multi-faceted evaluation approach, incorporating human evaluation, GPT-4o evaluation, and traditional metrics for a comprehensive analysis. For the human evaluation component, we have established a set of positive criteria that cover categories such as color, shape, function, material, relational position, spatial features, and geometric characteristics. Experts assign scores to the generated captions based on visualizations of the corresponding point cloud, rather than the captions produced by GPT-4 in the PointLLM setting. To minimize potential bias, multiple experts participate in the scoring process, and the results are reviewed and refined to ensure the scores provide a reliable standard for evaluating caption quality.

For GPT-4o evaluation, we compare the generated caption with the original ground truth caption. GPT-4o assesses how well the aspects highlighted in the original caption are reflected in the model's output, calculating the percentage of these aspects that are accurately or partially captured, with scores ranging from 0 to 100. When evaluated using traditional metrics such as BLEU-1 \cite{papineni2002bleu}, ROUGE-L \cite{lin2004rouge}, and METEOR \cite{banerjee2005meteor}, PointLLM-PiSA demonstrates strong performance. However, it is important to note that these metrics may have limitations, as discussed by PointLLM \cite{xu2023pointllm}, which some readers may find relevant. Despite these potential shortcomings, we provide results to highlight the effectiveness of PointLLM-PiSA. Additionally, we employ SentenceBERT \cite{reimers2019sentence} and SimCSE \cite{gao2021simcse} to measure the similarity between sentence embeddings of captions generated by PointLLM-PiSA and annotated references.

As shown in Table~\ref{tab:obj_caption}, PointLLM-PiSA sets a new \textit{state-of-the-art} on GPT-4o, human evaluation, and traditional metrics, achieving a notable improvement of \textbf{\textit{+16.46\%}} on GPT-4o. Additionally, the model demonstrates enhanced performance on BLEU-1, ROUGE-L, and METEOR. The performance of the next iteration also represents better performance. For detailed scoring criteria of the human evaluation, please refer to Appendix \emph{\textbf{C}}. 

In light of several identified errors in the ground truth of PointLLM, we aim to contribute to the advancement of the community by introducing PiSA-Bench. On PiSA-Bench, PointLLM-PiSA continues to lead in multiple evaluation criteria. As presented in Table~\ref{table:pisa-bench_cap}, PointLLM-PiSA-13B achieve \textit{state-of-the-art} performance, with significant improvements of 46.91\% \textbf{\textit{(+5.78\%)}}. 

Specifically, the ability of PointLLM-PiSA is continuously refined through our self-augmented training process. Remarkably, without any additional retraining, PointLLM-PiSA maintains generalizability to unseen objects, highlighting its robustness. Results indicate that iterative rounds of training provide consistent performance gains, with PointLLM-PiSA$^{2}$-7B achieving a noteworthy performance of 46.41\% \textbf{\textit{(+8.29\%)}}, the score of PointLLM-PiSA$^{3}$-7B arriving at 46.45\%.

\subsection{3D Shape Classification}
\begin{figure}[t!]
\centering
\includegraphics[width=0.48\textwidth]{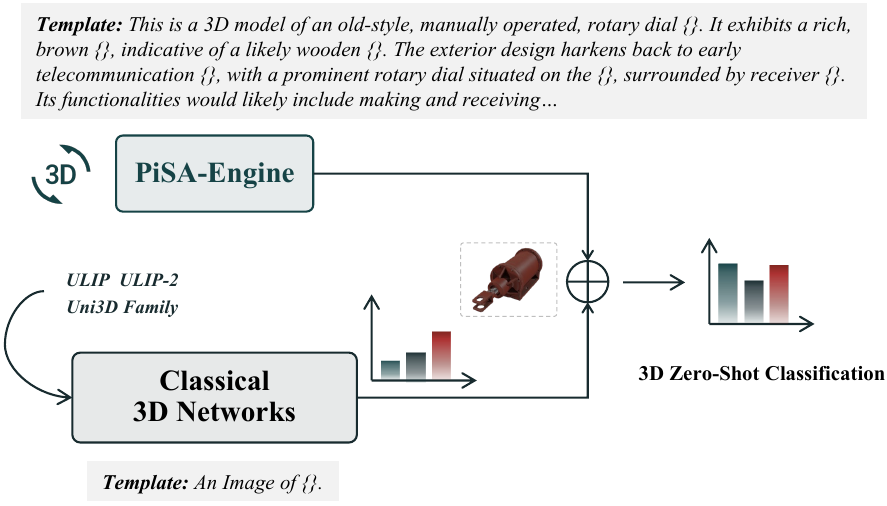}
\caption{\textbf{Plug-and-Play Module used for Downstream Tasks.}
The PiSA-Engine functions as a plug-and-play module that integrates 3D knowledge with the 2D knowledge, enhancing the model generalization ability.}
\label{PNP}
\end{figure}

PiSA-Engine can also be used for traditional 3D downstream tasks. Here, we leverage newly generated data to achieve \textit{state-of-the-art} performance in zero-shot 3D classification tasks. 

Specifically, we replace all core nouns in the data from PiSA-Engine with placeholders \{\} under the help of LLM, preserving only descriptive words. Thus, the prompt given to the LLM is, \textit{``Replace the core noun word in the sentence by \ \{ \} \ , while keeping the other information accurate."} This approach generates a 3D template derived from processed 3D captions, enabling the calculation of higher similarity to target labels, as shown in Figure~\ref{PNP}. When using ULIP as a baseline for zero-shot classification tasks, our method achieves significantly higher accuracy with an improvement of 3.56\%.

Additionally, we incorporate this method as a \textbf{\textit{plug-and-play enhancement module}} in Uni3D~\cite{zhou2023uni3d}, resulting in a 64.42\% accuracy, setting a new \textit{state-of-the-art} for zero-shot classification tasks on ScanObjectNN and even surpassing several fully trained methods. Our outstanding performance can be seen in Appendix \emph{\textbf{B}}.

\subsection{Ablation Study}
\label{ablation}
\begin{table}[t!]
\centering
\setlength\tabcolsep{10pt}
{\small
\begin{tabular}{@{}>{\centering\arraybackslash}p{6cm}c@{\hspace{1em}}p{1.25cm}@{}}
\toprule
Model & AVG \\ \midrule
\rowcolor{black!10}\multicolumn{2}{c}{\textit{w. 2D \& 3D knowledge}} \\ \midrule
PointLLM-7B & 38.12 \\
PointLLM-PiSA-7B \textit{(trained with 82K)} & 41.84 \\
\textbf{PointLLM-PiSA-7B }\textit{(trained with 132K)} & \textbf{43.70} \\
\midrule
\rowcolor{black!10}\multicolumn{2}{c}{\textit{w/o. 2D knowledge}} \\ \midrule
PointLLM-PiSA-7B & 38.61 \\ 
\midrule
\rowcolor{black!10}\multicolumn{2}{c}{\textit{w/o. 3D knowledge}} \\ \midrule
PointLLM-PiSA-7B & 38.77 \\ 
\bottomrule
\end{tabular}}
\caption{\textbf{Ablation Study.} Above shows the 3D object captioning results of PointLLM-PiSA on PiSA-Bench under different scaling training data and the data-filter model in PiSA-Engine. Considering the randomness of the generative task, we conduct five tests and take the average as the final result.}
\label{table:abla}
\end{table}

\paragraph{Data-filter Model.} 
In this study, we examine the role of different models as data filters within the PiSA-Engine. One approach utilizes 2D MLLMs like Qwen2-VL, which process 2D data generated from 12 rendered images of the point cloud. Alternatively, we consider LLMs like Gemma, which simplify data processing by operating without 2D image input and instead focus solely on refining semantic descriptions. The results, presented in Table~\ref{table:abla}, indicate that models incorporating 2D knowledge yield a substantial performance boost. We also exclude 3D knowledge and find that the results are inferior to our method.

\paragraph{Training Data.} We compared the model’s performance when trained on two datasets: one with 82K instructions (12K generated and 70K original) and another with 132K instructions (62K generated and 70K original). As shown in Table~\ref{table:abla}, the model trained on the larger dataset of 132K instructions demonstrates superior performance.

\section{Conclusion}
\label{conclusion}
Our research explores PiSA, a Self-Augmented Data Engine and Training Strategy for 3D Understanding with MLLMs.
PiSA-Engine enables a continuous cycle of high-quality data generation through three stages. Through this co-evolution training framework, 
PointLLM-PiSA shows \textit{state-of-the-art} results on GPT-4o, traditional, human evaluation, and achieves significant performance of 46.45\% \textbf{\textit{(+8.33\%)}} and 63.75\% \textbf{\textit{(+16.25\%)}} in zero-shot 3D object captioning and generative classification tasks on PiSA-Bench, our proposed comprehensive benchmark. Our method is designed as a plug-and-play framework to enhance the specific method and show extensive scalability. 

{
    \small
    \bibliographystyle{ieeenat_fullname}
    \bibliography{main}
}
\newpage
\setcounter{page}{1}

\clearpage
\begin{center}
    \textbf{\large Appendix Overview}
\end{center}

In the supplementary section, we first provide more related work and discussions to clarify existing solutions to self-augmented large language models. In addition, we provide more experimental details and results of PointLLM-PiSA. We organize our supplementary materials as follows:

\begin{itemize}
\item In ~\cref{sec:supp_dataset}, we provide more details of data components. \\
\item In ~\cref{supp:3d_zs}, we provide the results of 3D zero-shot classification task through the data processed by PiSA-Engine.\\
\item In ~\cref{sec:supp_vis}, we clarify the prompts in PiSA-Engine and PiSA-Bench, as well as the human scoring criteria.
\end{itemize}
\renewcommand{\thesection}{\Alph{section}}
\setcounter{section}{0}

\section{More Details on Dataset Construction}
\label{sec:supp_dataset}
Following the conventional approach for constructing instruction-following datasets, the original training data for PointLLM is generated by GPT-4. Specifically, GPT-4 is prompted to create a diverse range of instructions that align with captions described in the Cap3D~\cite{deitke2023objaverse} captions. This resulted in a large-scale dataset combining point-text instructions, comprising 660K brief descriptions and 70K detailed instructions.

For the training of PointLLM-PiSA, we extract 660K objects from the Cap3D~\cite{deitke2023objaverse} dataset. From this dataset, 3,000 objects are set aside exclusively for future testing, ensuring they are completely excluded from all stages of model training. Of these 3,000 reserved objects, 200 are used for testing in PointLLM, while the 40 objects in PiSA-Bench are also selected from this segregated set to prevent any information leakage.

To address the labor-intensive nature of manual data collection, we generate an additional 62K instructions using the PiSA-Engine. These newly created instructions were combined with the original 70K detailed instructions, resulting in a total of 132K complex instruction datasets for Stage 3.
As shown in Table~\ref{app_tab:general_statis_data}, despite our processed data accounting for only \textbf{13.28\%} of the overall training dataset for PointLLM-PiSA, it delivers a significant improvement in performance.

\section{3D Zero-shot Classification}
\label{supp:3d_zs}
We first obtain method ULIP~\cite{xue2022ulip} pre-trained from ShapeNet~\cite{chang2015shapenet} and the 3D-caption data processed by our PiSA-Enigne. As shown in Table~\ref{tab:ulip}, with templated-based prompt engineering, the accuracy of ULIP 59.56\% is enhanced to 63.12\% with a significant improvement of \textit{\textbf{+3.56\%}}.
\zl{Moreover, our PiSA-Enigne also enhances ULIP-2 to achieve better performance, with an accuracy of 72.93\% to 73.58\%.}

For Uni3D~\cite{zhou2023uni3d} pre-trained by the ensembled dataset, the \textit{state-of-the-art} on zero-shot classification, we ensemble two predicted logits (the original one and the one processed through our method) by simple addition as the final output.
Remarkably, with the integration of 3D-caption data processed by the PiSA-Engine, Uni3D-L of 58.24\% is enhanced to 59.15\% \textit{(Top-1)} with an improvement of \textit{\textbf{+0.91\%}} for the challenging ScanObjectNN dataset, as shown in Table~\ref{tab:uni3d_Scan}. This ensemble of two modalities highlights the complementary interactions between knowledge sources, enhancing overall model performance.

\begin{table}[t!]
\centering
\begin{adjustbox}{width=\linewidth}
\begin{tabular}{lcc}
\toprule
Number of all samples  &&\textit{Proportion (\%)} \\
\midrule
Total & 843,949 & 100 \\
Brief-description Type & 661,577 & \textit{78.39} \\
Detailed-description Type & 15,055 & \textit{1.78}\\
Single-round Type & 40,122 & \textit{4.75} \\
Multi-round Type & 15,097 & \textit{1.79} \\
\textbf{Instructions} \textit{(from PiSA-Engine)} & \textbf{112,098} & \textbf{\textit{13.28}} \\
\bottomrule
\end{tabular}
\end{adjustbox}
\caption{\textbf{Statistics of 3D Instruction Following Data.}}
\label{app_tab:general_statis_data}
\end{table}

We further applied this ensembling method to additional datasets and observed similar improvements in performance. These experimental results confirm that the proposed method can serve effectively as a \textbf{\textit{plug-and-play enhancement module}} for existing approaches, enabling robust point cloud understanding.
\begin{table}[t]
  \setlength{\tabcolsep}{2pt}
  \centering
  {\fontsize{9pt}{9pt}\selectfont
  \scalebox{0.95}{
    \begin{tabular}{c|cc}
    \toprule
    \multirow{1}[4]{*}{Method} & \multicolumn{2}{c}{\textbf{ModelNet40}} \\
  & Top1  & Top5 \\
    \midrule
    PointCLIP& 19.3  & 34.8   \\
    PointCLIPv2& 63.6  & 85.0   \\
    ReCon\cite{qi2023contrast}& 61.2  & 78.1   \\
    CG3D& 48.7 & 66.5 \\
    CLIP2Point& 49.5  & 81.2   \\
    \midrule

    ULIP& 59.56  & 83.95   \\
    \textbf{ULIP-\textit{PiSA-Engine}} & \cellcolor{gray!10}\textbf{63.12} \textcolor{DarkGreen}{($\uparrow 3.56$)} & \cellcolor{gray!10}\textbf{85.45} \textcolor{DarkGreen}{($\uparrow 1.50$)}\\
    ULIP-2  & 72.93   & 91.37  \\
    \textbf{ULIP-2-\textit{PiSA-Engine}} & \cellcolor{gray!10}\textbf{73.58} \textcolor{DarkGreen}{($\uparrow 1.65$)}          & \cellcolor{gray!10}\textbf{92.22} \textcolor{DarkGreen}{($\uparrow 0.85$)}  \\
    \bottomrule
    \end{tabular}}
    \par} 
\caption{\textbf{Zero-shot Classification on ModelNet40~\citep{wu20153d}.} \textit{(We retest the ULIP family via open-source code for the sake of rigor. 
)}}
\label{tab:ulip}
\end{table}

\begin{table}[t]
  \setlength{\tabcolsep}{2pt}
  \centering  
  {\fontsize{9pt}{10pt}\selectfont
  \scalebox{0.84}{
    \begin{tabular}{c|ccc}
    \toprule
    \multirow{2}[2]{*}{Method}   & \multicolumn{3}{c}{\textbf{ScanObjectNN}} \\
\cmidrule{2-4}        & Top1  & Top3  & Top5 \\
    \midrule
    OpenShape-SparseConv 	& 56.7 & 78.9 & 88.6\\
    OpenShape-PointBERT 	& 52.2 & 79.7 & 88.7\\
    ULIP-PointBERT    & 51.6 & 72.5 & 82.3\\
    \midrule
    Uni3D-Ti 	& 60.90	& 79.86	& 88.58 \\
    \textbf{Uni3D-Ti-\textit{PiSA-Engine}} & \cellcolor{gray!10}\textbf{61.49} \textcolor{DarkGreen}{\scriptsize ($\uparrow 0.56$)} & \cellcolor{gray!10}\textbf{82.04}\textcolor{DarkGreen}{\scriptsize ($\uparrow 2.18$)} & \cellcolor{gray!10}\textbf{89.72} \textcolor{DarkGreen}{\scriptsize ($\uparrow 1.14$)} \\	 
    Uni3D-B  & 63.88 & 82.73 & 90.27  \\
    \textbf{Uni3D-B-\textit{PiSA-Engine}} & \cellcolor{gray!10}\textbf{64.42} \textcolor{DarkGreen}{\scriptsize ($\uparrow 0.54$)} & \cellcolor{gray!10}\textbf{83.63} \textcolor{DarkGreen}{\scriptsize ($\uparrow 0.90$)}& \cellcolor{gray!10}\textbf{91.38} \textcolor{DarkGreen}{\scriptsize ($\uparrow 1.11$)} \\
    Uni3D-L & 58.24 & 81.81 & 89.41 \\
    \textbf{Uni3D-L-\textit{PiSA-Engine}} & \cellcolor{gray!10}\textbf{59.15}\textcolor{DarkGreen}{\scriptsize ($\uparrow 0.91$)}  & \cellcolor{gray!10}\textbf{83.37}\textcolor{DarkGreen}{\scriptsize ($\uparrow 1.56$)}  & \cellcolor{gray!10}\textbf{90.87}\textcolor{DarkGreen}{\scriptsize ($\uparrow 1.46$)}   \\
    \bottomrule
    \end{tabular}}
    \par} 
   \caption{\textbf{Zero-shot Classification on ScanObjectNN~\citep{uy-scanobjectnn-iccv19}.} \textit{(Uni3D-Ti, Uni3D-B, and Uni3D-L represent Uni3D's tiny, base, and large versions, respectively.)}} 
    \vspace{-0.5cm}
  \label{tab:uni3d_Scan}
\end{table}

\section{Additional Results}
\label{sec:supp_vis}
\subsection{Data Filter Prompts}
In the PiSA-Engine framework, 2D MLLMs such as Qwen2-VL~\cite{Qwen2VL} serve as supervisor, tasked with verifying and refining the responses generated by 3D MLLMs based on the provided images. This refinement process is detailed in Table~\ref{tab:qwen_prompt}, where the \textbf{\textit{Input}} corresponds to the 3D caption produced by the 3D MLLMs, and the \textbf{\textit{Output}} represents the enhanced response generated by Qwen2-VL.

\subsection{PiSA-Bench Evaluation}

\paragraph{Generative 3D Object Classification.}
In this task, GPT-4o acts as an evaluator to assess whether the model’s response aligns with the object type described by the class in PiSA-Bench. The procedure is outlined in Table~\ref{tab:open_vocabulary_gpt-4o_prompt}, where \{ground\_truth\} represents the PiSA-Bench, and \{model\_output\} refers to the response of the model. While the response of the model does not need to exactly replicate the ground truth class, it must correctly identify the object type.

\paragraph{3D Object Captioning.}
In this task, we utilize GPT-4o as an evaluator to compare captions generated by the model against PiSA-Bench, which serves as the ground truth. GPT-4o identifies elements from the PiSA-Bench captions and assesses the degree to which these elements are accurately or partially represented in the model-generated captions. Scoring is conducted on a scale of 0 to 100, with each identified element contributing equally to the overall score. The evaluation process is detailed in Table~\ref{tab:object_captioning_gpt-4o_prompt}, where the \{ground\_truth\} corresponds to the PiSA-Bench caption and the \{model\_output\} refers to the caption generated by the model. Furthermore, the six aspects used for evaluation are explicitly defined and explained to GPT-4o to enhance understanding and accuracy, as outlined in Table~\ref{tab:object_captioning_gpt-4o_prompt}. Table~\ref{tab:object_captioning_positive} and Table~\ref{tab:object_captioning_negative} present examples of positive and negative caption evaluations, respectively.

\subsection{Human Scoring Criteria}
Human evaluators are employed to assess captions in the object captioning task. To prevent bias, outputs corresponding to the same object from different models are grouped together and shuffled randomly. Evaluators then independently rate these captions while manually examining the colored visualization of the associated point cloud. Rather than employing captions generated by models as the ground truth, we rely on the native point cloud visualization to ensure more accurate and reliable evaluation results.
The evaluation process involves assigning correctness score following these guidelines:

\paragraph{Correctness Score.}
        Each distinct correct attribute in a model output is awarded one point, including category, color, shape, usage, material, relative position, spatial, and geometric information. For example, a white dinosaur correctly identified as a dinosaur and being white would receive two points.
        Partial correctness is graded on a scale of 0 to 1, depending on the degree of accuracy. For instance, if a model output described ``a creature resembling a lizard" but the object is specifically a dinosaur, it would be awarded 0.75 points.

\begin{table*}
\begin{tabular}{@{}l p{0.89\linewidth}@{}} 
\rowcolor{gray!12} & \\
\rowcolor{gray!12}\textbf{\textit{Prompt}} & 
\begin{minipage}[t]{0.99\linewidth}

You are a helpful AI assistant. Now I will give you a description of a 3D model and several rendering images of this 3D model. 
\\
\\
You should correct the incorrect content that does not match the images, and refine this 3D description according to the given images in terms of its appearance and details. 
\\
\\
Do not edit depth, spatial, or relative position information: \textcolor{DeeperBlue}{\{model\_output\}}
\\

\end{minipage} 
\\ \midrule \midrule
Sample 1 &
\begin{minipage}[t]{0.99\linewidth}
...\\
\textit{\textbf{Input:}} The 3D model represents a natural geological formation known as a cave. This cave is distinctly characterized by its many interconnected rooms and passageways, some of which are unoccupied. The cave's layout indicates several possible exploration paths, suggesting a level of intrigue and adventure often associated with spelunking activities. The cave model could be used in various digital scenarios such as gaming, film production, and geological studies, among others. \\
\textit{\textbf{Output:}} The 3D model represents a complex structure resembling a shipwreck. This shipwreck is characterized by its fragmented and damaged appearance, with various parts of the ship scattered and broken. The model includes elements such as wooden planks, metal structures, and debris, indicating a scene of destruction and abandonment. The shipwreck model could be used in various digital scenarios such as historical reconstructions, disaster simulations, or artistic projects. \\
\end{minipage} \\ \midrule
Sample 2 &
\begin{minipage}[t]{0.99\linewidth}
...\\
\textit{\textbf{Input:}} This is a model of a split air conditioner unit's top part, primarily coloured in white with a dark grey rectangular panel at the top. It is designed to expel heat from the conditioned space, making it essential for cooling purposes. The dark grey panel possibly indicates the unit's control panel or display unit. The model reflects a common type of split air conditioner, often used in home and office spaces for temperature control. \\
\textit{\textbf{Output:}} This is a model of a window air conditioning unit, primarily colored in white. The unit features a rectangular shape with a front grille that has horizontal slats for air intake and exhaust. There is a small control panel on the right side of the front grille, which likely includes buttons for operation and a digital display. The model represents a common type of window-mounted air conditioner, widely used in residential and commercial settings for cooling individual rooms. The design suggests functionality and ease of installation in windows. \\
\end{minipage} \\ \midrule
Sample 3 &
\begin{minipage}[t]{0.99\linewidth}
...\\
\textit{\textbf{Input:}} This 3D model depicts a vibrantly animated, multi-colored train in motion. The train's various parts - including the engine, carriages, and wheels - are painted with an array of colors, giving it a playful, kid-friendly aesthetic. The movement of the train suggests that it's either pulling or pushing a number of passengers, indicating its use in passenger transportation, commonly seen in railway systems worldwide. \\
\textit{\textbf{Output:}} This 3D model depicts a single, stationary train carriage painted in a combination of red and yellow. The train has multiple windows along its sides and is designed for passenger transportation, as indicated by the number ``2" on both ends of the carriage. The model does not show any movement or additional carriages, suggesting it represents a single unit rather than a full train in motion. \\
\end{minipage} \\

\bottomrule
\end{tabular}
\caption{\textbf{The Prompt and Samples of Data Filter in PiSA-Engine.} The data filter is told to refine the visual details of 3D caption while not to amend the 3D information. The placeholder \textcolor{DeeperBlue}{\{model\_output\}} refers to the 3D caption.
}
\label{tab:qwen_prompt}
\end{table*}

\begin{table*}
\begin{tabular}{@{}l p{0.89\linewidth}@{}} 
\rowcolor{gray!12} & \\
\rowcolor{gray!12}\textbf{\textit{Prompt}} & 
\begin{minipage}[t]{0.99\linewidth}

You are a helpful AI assistant. Now, I will give you an answer from the model and an answer from the label.\\
\\
All you need to do is focus on these two answers and figure out whether they are referring to the same general class, 
focusing on the class of object, not attributes such as color, shape, count, spatial or usage.
Respond with `T' if they refer to the same class and `F' if not. Also, provide a brief rationale (no more than 20 words) for your judgment.\\
\\
Remember, the answer from the model refers to one of the answers from the label; even if the answer from the model refers to the subclass of one of the answers from the label, you should respond as `T'.
Your response should follow the judgement standard of the prompt I give.
Firstly, I will give you two examples of answer pairs as well as their responses:\\
\\
Example1:\\
Input: 1. wooden board, table, pottery 2. This is a 3D model of wooden table. \\
Output: T\#Both refer to a table.\\
\\
Example2:\\
Input: 1. historical vehicle, pioneer wagon, covered wagon, prairie schooner 2. The 3D object model depicts a quaint, old-fashioned cart. The cart is entirely brown, with two sturdy wooden wheels for mobility. The main body of the cart is in the shape of a large, semicircular curve, made of wood and affixed to the wheels. This curved body extends backward, forming a simple, straight tail. Despite its simplicity, it reflects a nostalgic charm and could be used in settings like historical reenactments, antiquated transportation exhibits, or in visual media for a touch of old-world atmosphere. \\
Output: F\#One refers to a wagon, the other to a truck.\\
\\
Now, analyze the following:\\
\textit{\textbf{Input:}} 1. \textcolor{DeeperBlue}{\{ground\_truth\}} 2. \textcolor{DeeperBlue}{\{model\_output\}} \\
\textit{\textbf{Output:}} 
\\
\end{minipage} \\ \midrule \midrule
Sample 1 &
\begin{minipage}[t]{0.99\linewidth}
...\\
\textit{\textbf{Input:}} 1. awning, shelter, canopy, shade 2. This is a 3D model of a structure featuring a vibrant blue and white striped shade covering a rectangular bench underneath. The shade, with its alternating blue and white stripes, provides ample protection from the sun, hinting at its use in creating a comfortable, outdoor sitting area. The bench's sturdy structure implies it can support significant weight, making it a practical choice for both indoor and outdoor setups, such as in a patio, backyard, or even a poolside area. \\
\textit{\textbf{Output:}} \textcolor{DarkGreen}{T\#Both refer to a shade or shelter structure. \checkmark}
\end{minipage} \\ \midrule
Sample 2 &
\begin{minipage}[t]{0.99\linewidth}
...\\
\textit{\textbf{Input:}} 1. space exploration vehicle, Mars rover, robotic rover, planetary exploration equipment. 2. The 3D object model is a sleek, black robotic cleaning machine. It is designed with a compact and modern structure, indicating its ability to navigate through tight spaces. This machine likely utilizes advanced technology to autonomously clean floors, carpets, or other surfaces. Its primary function is likely to eliminate manual labor and time spent on cleaning, providing a deep clean with precision and ease. This can be used in both domestic and commercial environments, making it a convenient and efficient tool for maintaining cleanliness. \\
\textit{\textbf{Output:}} \textcolor{Darkred}{F\#One refers to a space exploration rover, the other to a robotic cleaning machine. \ding{55}}
\end{minipage} \\ 

\bottomrule
\end{tabular}
\caption{\textbf{The Prompt and Samples of GPT-4o in Open-vocabulary Classification.} GPT-4o needs to analyze two sentences to determine if they refer to the same general object or concept, focusing on the type of object, not attributes such as color, size, or shape. The placeholders \textcolor{DeeperBlue}{\{ground\_truth\}} and \textcolor{DeeperBlue}{\{model\_output\}} refer to the PiSA-Bench and the response of model, respectively.}
\label{tab:open_vocabulary_gpt-4o_prompt}
\end{table*}

\begin{table*}
\scalebox{0.95}{
\begin{tabular}{@{}l p{0.96\linewidth}@{}} 
\rowcolor{gray!12} & \\
\rowcolor{gray!12}\textbf{\textit{Prompt}} & 
\begin{minipage}[t]{0.99\linewidth}
You are a helpful AI assistant. Now, I will give you an answer from the model and an answer from the label.
All you need to do is to evaluate these two answers from six aspects separately: \\
1. ``description": Giving a comprehensive description of the whole 3D model.\\
2. ``color": Demonstrating the color attribute of the whole or the individual objects.\\
3. ``shape": Demonstrating the geometric attribute of the whole or the individual objects.\\
4. ``count": Counting the number of the whole or the individual objects.\\
5. ``spatial": Understanding the spatial relations between multiple objects in the 3D model.\\
6. ``usage": Demonstrating the usage or the production purpose of the 3D model.\\
\\
For any aspect above, you should identify the aspects mentioned in the answer from the model and calculate the percentage of these aspects correctly mentioned or partially matched in the answer from the label. 
Remember the score is to evaluate how much the two answers match.
When evaluating and comparing each criterion, do not take other criteria into consideration.
Score from 0 to 100. Consider similar concepts and synonyms. 

Your response should include the scores of the six criteria (description, color, shape, count, spatial, and usage score) in the order above.
Remember all scores range from 0 to 100. 
Firstly, I will give you several answer pairs and their corresponding scores. Your response format should follow the example of the prompt I give:\\
\\
Provide your score (0-100) in the format of below:\\
`
Scores for each aspects: **[description score, color score, shape score, count score, spatial score, usage score]**
'\\
For clarity, consider this example:\\
\\
\textbf{\textit{Label:}} \\
``description": ``This 3D model displays a wooden rectangular platform adorned with various items on top. The setup features a sizable dark clay vessel or cauldron positioned on a miniature stand at one end of the platform. At the other end, a table-like structure supports books, miniature vases, and what seems to be a witch's hat. The entire arrangement is decorated with vibrant speckles or splashes of paint in red, green, and blue, creating a magical or whimsical feel. The wood surface shows clear plank marks and a textured finish.",\\
``color": ``The base displays a wooden brown shade, decorated with vibrant speckles in red, green, and blue. The large vessel is dark gray. The table holds brown books, two small vases (one blue and one green), and a dark gray pointed hat. The table also features the same speckled pattern as the wooden platform.",\\
``shape": ``The principal structure is a rectangular wooden platform. It includes a circular pot at one end and a rectangular table at the other.",\\
``count": ``The arrangement includes several items: one large pot, one table, multiple books, at least two small vases, and one pointed hat.",\\
``spatial": ``The clay vessel is situated at one end of the wooden platform. The table, bearing various items, is located at the opposite end. All elements are positioned on the wooden platform.",\\
``usage": ``This 3D model can be used in video games, animations, or virtual settings to craft a scene with a magical or fantasy theme, perhaps associated with witchcraft or alchemy. It can also be used for reading, writing, and eating."\\
\\
\textbf{\textit{Model:}} ``This model portrays a vivid scene of a cartoon-styled table, overflowing with a variety of objects. The table seems to be in use, showcasing a naturalistic depiction of a cluttered table in a domestic or workspace. Objects on it reflect common items like books, pencils, cups, etc. suggesting its functionality as a piece of furniture where different activities such as reading, writing, or drinking can be performed."\\
\\
\textbf{\textit{Output:}} 
Scores for each aspects: **[35,0,0,30,35,75]**\\
\\
Now score the following:\\
\textit{\textbf{Label:}} \textcolor{DeeperBlue}{\{ground\_truth\}} \\
\textit{\textbf{Model:}} \textcolor{DeeperBlue}{\{model\_output\}} \\
\textit{\textbf{Output:}} \\
\end{minipage} 
\\ \bottomrule
\end{tabular}}
\caption{\textbf{The Prompt of PiSA-Bench in Object Captioning.} GPT-4o evaluates the response of the model by identifying aspects mentioned in the PiSA-Bench caption and calculating the percentage of aspects that are correctly or partially matched in the caption generated by the model. The placeholders \textcolor{DeeperBlue}{\{ground\_truth\}} and \textcolor{DeeperBlue}{\{model\_output\}} refer to the PiSA-Bench caption and the response of the model, respectively.}
\label{tab:object_captioning_gpt-4o_prompt}
\end{table*}

\begin{table*}

\scalebox{0.9}{
\begin{tabular}{@{}l p{0.99\linewidth}@{}} 
\toprule
\\
Sample 1 & 
  \begin{minipage}{\linewidth}    
  \hspace{2em}
  \includegraphics[width=0.1\linewidth,height=0.2\linewidth,valign=c]{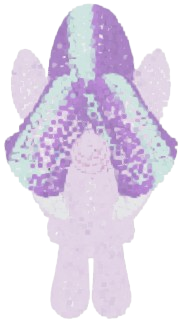}
  \hspace{3em}
  \includegraphics[width=0.2\linewidth,height=0.2\linewidth,valign=c]{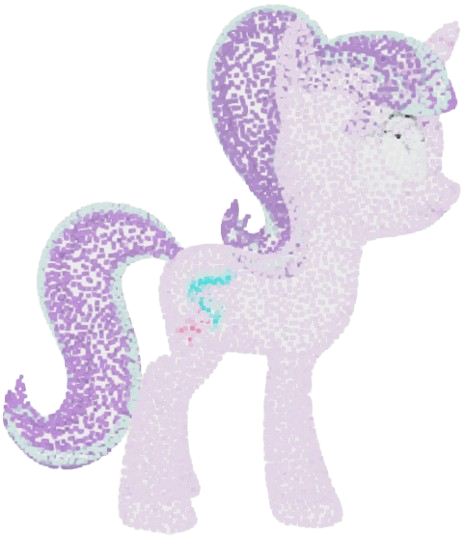}
  \hspace{2em}
  \includegraphics[width=0.2\linewidth,height=0.2\linewidth,valign=c]{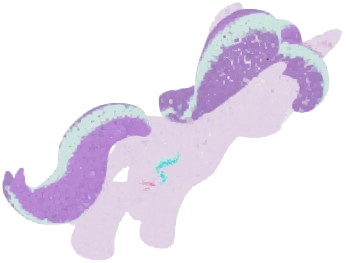}
  \hspace{3em}
  \includegraphics[width=0.1\linewidth,height=0.2\linewidth,valign=c]{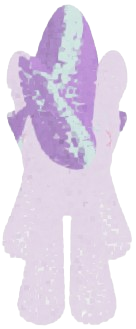}

  \hspace{0.8em}
\end{minipage}
\\ \\ \midrule
 & \begin{minipage}[t]{0.99\linewidth}
...\\
\textit{\textbf{Ground Truth:}} ``description": ``This 3D model represents a whimsical unicorn pony, styled like those seen in `My Little Pony'. It sports a light lavender body complemented by a purple mane and tail. The unicorn is adorned with a solitary horn and large, vivid eyes. The mane and tail display a dual-color design, predominantly purple with a stripe of mint green. On its side, it bears a small turquoise emblem or `cutie mark'. The overall design of the model is smooth and rounded, typical of cartoon figures intended for a young audience.", ``color": ``The body is shaded light lavender. The mane and tail are mostly purple, accented with mint green stripes. The horn is pale pink. A small turquoise emblem is present on its side. The eyes are plain white, lacking iris or pupil in these images.", ``shape": ``The pony features a stylized, rounded body with slender legs, a petite snout, and an oversized head, typical of cartoon creatures. The mane and tail are designed with a flowing, wavy form. The horn exhibits a twisted shape.", ``count": ``There is one pony.", ``spatial": ``", ``usage": ``This 3D model is ideal for use in children's animated TV shows or movies, especially those akin to the `My Little Pony' series. It could also be utilized in video games, mobile applications, or as a prototype for toy designs. Additionally, the model is suitable for digital artwork, 3D animation guides, or as a part of a character collection in a larger animated setting. \\
\\
\textit{\textbf{Model:}} A 3D model of a charming, small cartoon pony. The pony is primarily beige in color, with vibrant purple hair in its mane and tail. Its eyes are large and expressive, being white, which amplifies its cartoonish look. The size and design suggest it could be a character from a kid's show or an animated series, usually used for entertainment purposes like TV shows, toys, or animations. \\
\\
\textit{\textbf{Output:}} \textcolor{DarkGreen}{Scores for each aspect: **[60, 50, 75, 100, 0, 85]**\\
1. **Description**: 60 - The model captures the general idea of a cartoon pony with a colorful mane and tail, but misses details like the horn and cutie mark, and does not specify My Little Pony styling.\\ 
2. **Color**: 50 - The model identifies the purple hair correctly but describes the body color as beige instead of light lavender, and misses mint green stripes in the mane and the turquoise emblem. \\
3. **Shape**: 75 - The rounded body, large eyes, and cartoonish style are captured well, but there is no mention of the horn or the characteristic slender legs and snout described in the label. \\
4. **Count**: 100 - Both answers agree there is one pony. \\
5. **Spatial**: 0 - The model does not include any spatial details. \\
6. **Usage**: 85 - The model accurately describes the usage similar to the label but lacks specificity regarding My Little Pony series and other possible uses like digital artwork."}\\
\\
\end{minipage}
\\ \bottomrule
\end{tabular}}
\caption{\textbf{The Positive Sample of PiSA-Bench in Zero-shot Object Captioning.} \textit{(The ``spatial" aspect being null indicates that the 3D object is a single, cohesive entity, without any spatial relationships or information about separate components.)}}
\label{tab:object_captioning_positive}
\end{table*}

\begin{table*}

\scalebox{0.9}{
\begin{tabular}{@{}l p{0.99\linewidth}@{}} 
\toprule
\\
Sample 2 & 
  \begin{minipage}{\linewidth}    
  \hspace{2em}
  \includegraphics[width=0.15\linewidth,height=0.2\linewidth,valign=c]{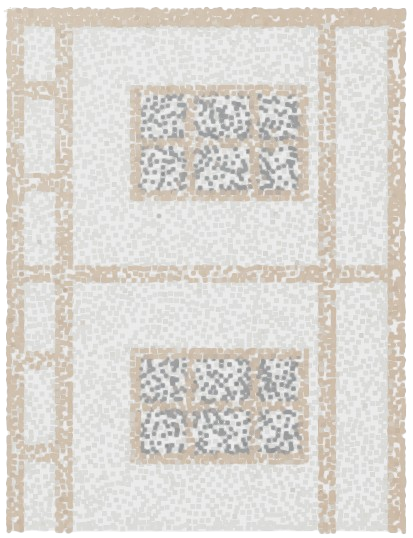}
  \hspace{3.5em}
  \includegraphics[width=0.15\linewidth,height=0.2\linewidth,valign=c]{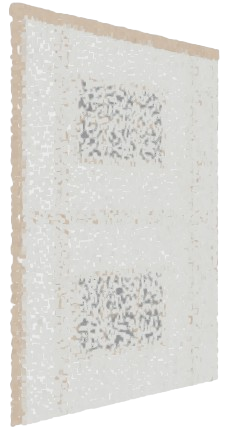}
  \hspace{3em}
  \includegraphics[width=0.15\linewidth,height=0.2\linewidth,valign=c]{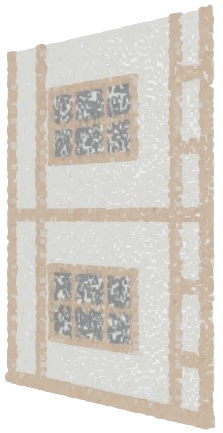}
  \hspace{4em}
  \includegraphics[width=0.02\linewidth,height=0.2\linewidth,valign=c]{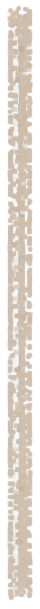}
  \hspace{0.8em}
\end{minipage}
\\ \\ \midrule
 & \begin{minipage}[t]{0.99\linewidth}
...\\
\textit{\textbf{Ground Truth:}} ``description": ``This 3D object depicts a portion of a building wall, equipped with two windows. The color of the wall ranges from light blue to off-white, complemented by window frames and trim in shades of tan or light brown. Each window is divided into six panes configured in a 2x3 layout. Above these, a series of smaller, rectangular transom windows are placed. The model includes intricate details like the framing around the windows, paneling on the walls, and ornamental trim.'', ``color": ``The primary wall color is either light blue or off-white, while the detailing around the windows and other architectural features are in tan or light brown. The glass of the window panes is a darker blue-gray.'', ``shape": ``The model is shaped as a rectangle, featuring two main rectangular windows. It incorporates various geometric forms, such as the square and rectangular panes and the straight lines of the trim.'', ``count": ``The model contains two primary windows, each with six panes, and six smaller transom windows positioned above the main ones.'', ``spatial": ``'', ``usage": ``This 3D model is suitable for use in architectural visualizations, interior design applications, video game settings that include building interiors or exteriors, or as a guide for building or remodeling projects.''\\
\\
\textit{\textbf{Model:}} A rectangular flat board featuring a light hue, with a smooth texture. It is outfitted with a noticeable wooden structure, comprising of two vertical posts and two horizontal beams, creating a 'U' shape. This kind of object could be a component of a furniture item, like a cabinet or wardrobe, with the wooden structure providing support and organization benefits.\\
\\
\textit{\textbf{Output:}} \textcolor{Darkred}{Scores for each aspects: **[20, 25, 25, 20, 0, 30]** \\
1. **Description**: 20 - The model describes a rectangular board with a 'U' shape wooden structure, which aligns partially with the label's discussion of a wall section equipped with windows. However, the key architectural elements (windows, panels, ornamental trim) are missing.\\
2. **Color**: 25 - The model mentions a ``light hue", partially matching the label's light blue/off-white description, but lacks specific details on other color features like tan/light brown window trim.\\
3. **Shape**: 25 - Both refer to a rectangular form, and there's mention of structures (the `U' shape in the model); however, the detailed shapes of window panes and trims described in the label are not addressed.\\
4. **Count**: 20 - The model does not address the count explicitly but there is a mention of ``two vertical posts and two horizontal beams", which could partially align with counting window frames but misses the specific panes.\\
5. **Spatial**: 0 - The model does not describe any spatial relations, whereas the label doesn't include spatial relationships either.\\
6. **Usage**: 30 - The model proposes usage related to furniture components, which is different from the architectural visualization purpose but relates to an object-based interpretation.}\\
\\
\end{minipage} \\ 
\bottomrule
\end{tabular}}
\caption{\textbf{The Negative Sample of PiSA-Bench in Zero-shot Object Captioning.} \textit{(The ``spatial" aspect being null indicates that the 3D object is a single, cohesive entity, without any spatial relationships or information about separate components.)}}
\label{tab:object_captioning_negative}
\end{table*}

\clearpage

\end{document}